\documentclass{article}

\usepackage[preprint, nonatbib]{neurips_2024}

\usepackage{multirow}
\usepackage[utf8]{inputenc} 
\usepackage[T1]{fontenc}    
\usepackage{hyperref}       
\usepackage{url}            
\usepackage{booktabs}       
\usepackage{amsfonts}       
\usepackage{nicefrac}       
\usepackage{microtype}      
\usepackage{xcolor}         

\usepackage{pifont}
\usepackage{enumitem}

\usepackage{graphicx}

\title{Instruct Large Language Models to Drive like Humans}

\author{
    \vspace{-5mm}
     \\
    Ruijun Zhang\textsuperscript{1$^*$},
    Xianda Guo\textsuperscript{2$^*$$^{\dag}$},
    Wenzhao Zheng\textsuperscript{3$^*$},
    Chenming Zhang\textsuperscript{4},\\
    Kurt Keutzer\textsuperscript{3},
    Long Chen\textsuperscript{1,4,5$^{\ddag}$} \\
    \textsuperscript{1} Institute of Automation, Chinese Academy of Sciences\\
    \textsuperscript{2} School of Computer Science, Wuhan University\\
    \textsuperscript{3} University of California, Berkeley~~~~~~\textsuperscript{5} Waytous \\
    \textsuperscript{4} Institute of Artificial Intelligence and Robotics, Xi'an Jiaotong University
}

\begin{document}

\maketitle

\renewcommand{\thefootnote}{\fnsymbol{footnote}}
\footnotetext[1]{These authors contributed to the work equally.}
\footnotetext[2]{Project leader}
\footnotetext[3]{Corresponding authors:long.chen@ia.ac.cn}

\begin{abstract}
Motion planning in complex scenarios is the core challenge in autonomous driving. 
Conventional methods apply predefined rules or learn from driving data to plan the future trajectory.
Recent methods seek the knowledge preserved in large language models (LLMs) and apply them in the driving scenarios.
Despite the promising results, it is still unclear whether the LLM learns the underlying human logic to drive.
In this paper, we propose an InstructDriver method to transform LLM into a motion planner with explicit instruction tuning to align its behavior with humans. 
We derive driving instruction data based on human logic (e.g., do not cause collisions) and traffic rules (e.g., proceed only when green lights).
We then employ an interpretable InstructChain module to further reason the final planning reflecting the instructions.
Our InstructDriver allows the injection of human rules and learning from driving data, enabling both interpretability and data scalability.
Different from existing methods that experimented on closed-loop or simulated settings, we adopt the real-world closed-loop motion planning nuPlan benchmark for better evaluation.
InstructDriver demonstrates the effectiveness of the LLM planner in a real-world closed-loop setting.
Our code is publicly available at \url{https://github.com/bonbon-rj/InstructDriver}.

\end{abstract}

\section{Introduction}
\label{sec:introduction}

Autonomous driving technology is crucial for enhancing road safety, which can reduce traffic congestion and improve transportation efficiency. The widely adopted pipeline of autonomous driving encompasses perception~\cite{zhang2023completionformer, guo2023openstereo, duan2023diffusiondepth, duan2023diffusiondepth, bevformer, beverse, tpvformer, surroundocc, guo2023simple}, prediction~\cite{Dauner2023CORL,cheng2022gpir,cheng2023forecast}, and motion planning~\cite{hu2021fiery,gu2022vip3d,liang2020pnpnet,zheng2024genad}, which together enable a vehicle to navigate complex environments. Among these, motion planning is particularly important as it ensures the vehicle can move smoothly and safely by determining the optimal path and speed while avoiding collision.

Conventional rule-based methods utilize predefined rules and logical conditions for planning~\cite{idm, rule_based_method1, rule_based_method2}, which offers a high degree of interpretability yet struggles to account for all possible scenarios. Learning-based methods train on extensive autonomous driving scenario data, enabling models to learn and comprehend various scenarios~\cite{learning_based_method1, learning_based_method2, learning_based_method3, learning_based_method4, learning_based_method5, learning_based_method_mp3, urban_driver, plancnn, plantf}.

Despite the strong performance, they treat motion planning as a black-box prediction problem, raising concerns about whether the output trajectories align with human driving behaviors.
With the emergence of Large Language Models (LLMs), recent methods seek to transfer their knowledge to motion planning~\cite{lmdrive, drivingllm, gpt_driver, languagempc}.

Despite the encouraging results, they usually rely on predefined planning objectives and it remains unclear whether LLMs have indeed learned the underlying logic of human driving. 
The absence of real-world closed-loop evaluation further raises concerns about whether they can make planning decisions based on environmental data~\cite{nuscenes_limit1, nuscenes_limit2}.

\begin{figure}[tb]
  \centering
  \includegraphics[width=1.0\linewidth]{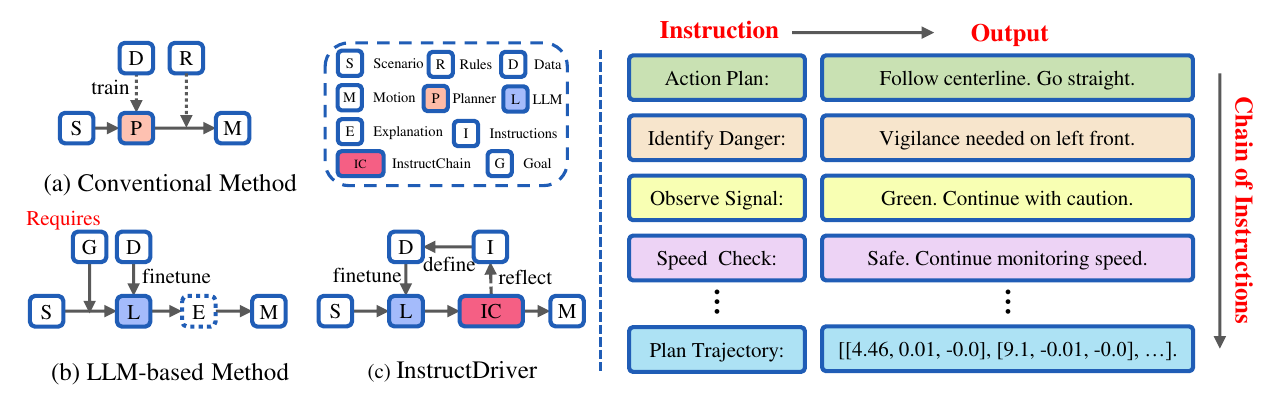}
  \caption
  {
  The motivation of our InstructDriver. The left figure compares different motion planning methods for autonomous driving, showcasing our method's ability to function without predefined objectives. It highlights how our method guides the planner to produce human-like driving behavior. The right figure illustrates the correspondence between the provided instructions and the resulting outputs.
  }
  \vspace{-0.4cm}
  \label{fig:introduction}
\end{figure}

To address this, we propose an InstructDriver method to align LLM-based planners with human behavior by generating a series of instructions based on the human logic of driving, as shown in Figure~\ref{fig:introduction}.
We further propose an InstructChain module to combine the instructions to reason about the final planning trajectory.
InstructDriver allows the incorporation of human rules and learns from driving data, thereby achieving both interpretability and data scalability.
By leveraging a sequence of intermediate instructions, InstructChain enhances the capability of LLM to handle the complex reasoning task of planning.

Our contributions are summarized as follows:
\begin{itemize}[leftmargin=0.5cm]
\item[$\bullet$] We introduce InstructDriver to align LLMs with a series of human instructions, ensuring their consistency with human logic for driving.
\item[$\bullet$] We propose InstructChain to enable LLMs to explicitly follow the execution of instructions, providing a high degree of interpretability.
\item[$\bullet$] We conducted extensive open-loop and closed-loop experiments within the nuPlan framework to validate the effectiveness of the proposed methods, achieving competitive performance metrics.
\end{itemize}

\section{Related Work}
\label{sec:related_work}

\textbf{LLM agents in planning tasks.} 
Agents implemented through LLMs have demonstrated exceptional proficiency in planning across various tasks and domains. For instance, LLMs have been utilized to develop a virtual town comprising 25 memory-capable agents \cite{virtual_town}, demonstrating their ability to reflect, plan, and make decisions in virtual environments, thus addressing complex communication tasks. Additionally, LLMs have been utilized to create an automated agent for Minecraft, which for the first time, successfully acquired all items in the Overworld technological tree \cite{minecraft_agents_ghost}. This achievement demonstrates LLMs' task-planning capabilities to navigate and accomplish complex objectives in intricate game settings. Moreover, LLMs have been used to develop frameworks for multi-agent collaboration, enabling multiple agents to communicate and cooperate effectively to complete tasks \cite{multi_agents_cooperate}. This highlights the capabilities of LLMs in multi-agent systems, demonstrating their planning abilities to enhance coordination and cooperation among agents. Furthermore, LLMs can be fine-tuned to generalizable, semantically aware robotics policies, used directly for robotic control \cite{RT-2}. This suggests that LLMs are not only applicable in virtual environments but also in real-world robotic control tasks. Agents implemented through LLMs possess reasoning and planning abilities, presenting promising prospects for motion planning in autonomous driving.

\textbf{Motion planner in autonomous driving without LLM.} 
The existing methods for motion planning without LLM can be broadly classified into three main types: rule-based, learning-based, and hybrid methods that integrate both. The rule-based method \cite{idm, rule_based_method1, rule_based_method2} represents a conventional approach, employing pre-established rules and logical frameworks to construct driving trajectories. Intelligent Driver Model (IDM) \cite{idm} is a heuristic motion model used to track the vehicle ahead in traffic while ensuring a safe following distance. With the development of deep neural networks, many learning-based methods \cite{learning_based_method1, learning_based_method2, learning_based_method3, learning_based_method4, learning_based_method5, learning_based_method_mp3, urban_driver, plancnn, plantf} have been proposed to handle complex driving scenarios by learning from extensive datasets of human driving behavior. For example, urban traffic data such as traffic lights, other vehicles, and pedestrians are used for training to learn rich urban driving strategies \cite{urban_driver}. PlanTF \cite{plantf}, a baseline model utilizing the Transformer architecture, focuses on the current state of autonomous vehicles rather than historical data, demonstrating superior generalization capabilities in long-tail scenarios. In the realm of hybrid methods, the PDM \cite{PDM} is highly distinguished. It effectively integrates rule-based approaches with learning-based methods, achieving competitive results. Currently, learning-based methods are considered as a potentially scalable solution for motion planning that can replace rule-based planners \cite{plantf}.

\textbf{LLM planner in autonomous driving.}
Recently, methods have been proposed to use LLM for motion planning in autonomous driving tasks. LanguageMPC \cite{languagempc}, by designing cognitive pathways to enable comprehensive reasoning in LLM and translating its decisions into actionable driving commands. GPT-Driver \cite{gpt_driver}, a GPT-based motion planner, represents the inputs and outputs of the planner as linguistic tokens and utilizes the LLM to generate driving trajectories via linguistic descriptions of coordinate positions. Driving-with-LLMs \cite{drivingllm} proposes an architecture at the object level that integrates vectorized numerical modalities with a pre-trained LLM to enhance the comprehension of context in driving scenarios. LMDrive \cite{lmdrive} is the pioneering work that utilizes LLM for closed-loop end-to-end autonomous driving. LLM-ASSIST \cite{llmassist} employs LLMs as an adaptive component within the other planner to enhance the performance of closed-loop planning. Utilizing LLM for motion planning improves adaptability to complex and dynamic road conditions, enhances decision-making interpretability, and boosts safety.

\section{Proposed Approach}
\label{sec:method}
\subsection{Motion Planning as Language Modeling}
\label{sec:motion_planning_as}
The objective of motion planning in autonomous driving is to design a safe and comfortable driving trajectory, denoted as $\mathcal{T}$, utilizing the observation data $\mathcal{O}$ and the current self-state $\mathcal{S}_t$ as inputs. More specifically, the inputs for motion planning should also encompass descriptions of the motion system  $\mathcal{S}_y$ to more accurately reflect system planning goals and the characteristics of the vehicle. Additionally, incorporating specific instructions $\mathcal{I}$ as input can effectively guide the planner to perform plans aligned with humans. This comprehensive set of information, when combined, provides a thorough basis for motion planning and can be represented textually as $X_t$:
\begin{equation}
X_{t} = \{\mathcal{O},\mathcal{S}_t,\mathcal{S}_y,\mathcal{I}\}.
\end{equation}
Furthermore, the output should not be limited solely to the motion trajectory $\mathcal{T}$ but should also encompass a thought chain for the autonomous driving decisions made during the planning process, referred to as InstructChain and denoted as $\mathcal{I}_c$. This inclusion ensures that the planning process remains transparent and interpretable:
\begin{equation}
Y_t = \{\mathcal{I}_c, \mathcal{T}\}.
\end{equation}
Through a process of logical transformation $F$, given the input $X_t$, the logarithm of the probability of the occurrence of each word in the corresponding output vocabulary can be obtained. This log-probability reflects the model's confidence in its predictions for each word, providing a quantified evaluation of the vocabulary distribution, as expressed in the following equation:
\begin{equation}
Y_{log} = F(X_t).
\end{equation}
To derive the final output $Y_t$ from the generated logical logarithm $Y_{log}$, a sequence of transformations is employed. This process involves temperature scaling $T$, which adjusts the distribution of the logits to control the randomness of the predictions. Subsequently, the Softmax function $S$ is applied to convert the scaled logits into probabilities. Finally, the top-p sampling strategy $P$ is utilized to select the most probable tokens:
\begin{equation}
Y_t = P(S(T(Y_{log},t)),p),
\end{equation}
where variable $t$ serves as the parameter for $T$, with higher values of $t$ correlating to greater diversity in the output. The variable $p$ is a parameter for top-p sampling, representing the probability mass.

In reference to the GPT-Driver \cite{gpt_driver}, by optimizing the logistic log $Y_{log}$ associated with the human InstructChain $\hat{\mathcal{I}_c}$ and the human driving trajectory $\hat{\mathcal{T}}$, the motion planner is capable of generating explanations and trajectories that closely resemble human driving behavior.

\begin{figure}[tb]
  \centering
  \includegraphics[width=1.0\linewidth]{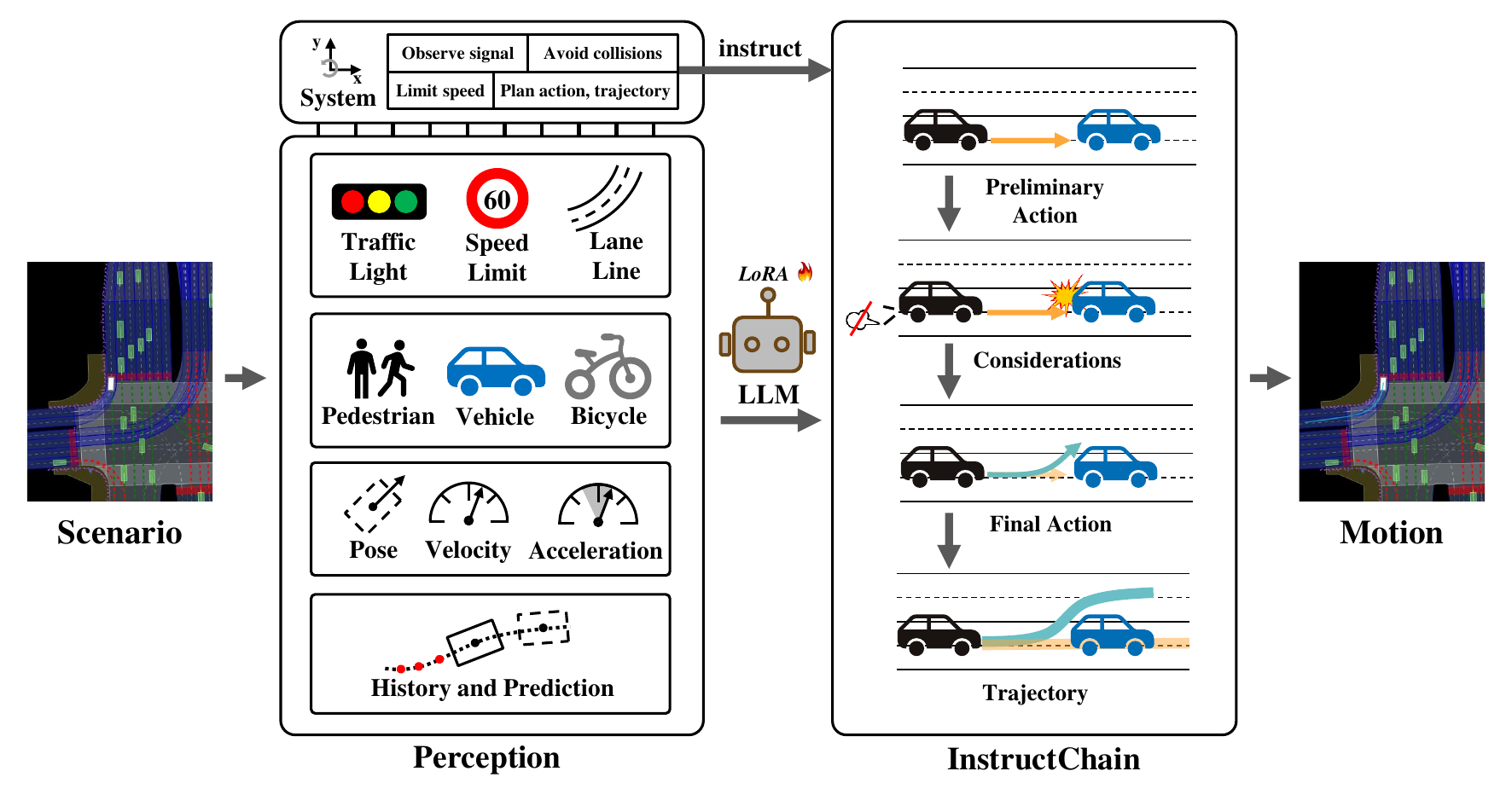}
  \caption
  {
  Overview of the motion planning process of our method InstructDriver. Our approach transforms scenario data into textual descriptions and, by setting specific instructions, enables a fine-tuned LLM to generate InstructChain and trajectories that align with human driving behavior. The trajectory is subsequently applied in a simulated environment.
  }
  \vspace{-0.4cm}
  \label{fig:overview}
\end{figure}

\subsection{Instruction-based Behavior Alignment} 
\label{sec:instruction_based_behavior_alignment} 
Through the aforementioned modeling process, it is necessary to construct $X_t$ and $Y_t$ for human drivers to perform imitation learning, thereby developing an autonomous driving model that emulates human behavior. This procedure involves a prompt design process, which we have named InstructDriver. The overall process of InstructDriver is illustrated in Figure \ref{fig:overview}. Initially, the scenario data is represented in natural language as $X_t$, encompassing both the instruction $I_t$ and the input $I_p$, which can be expressed as follows:
\begin{equation}
X_t = \{I_t, I_p\}.
\end{equation}
The instruction $I_t$ is used in the model to direct the decision-making process and overall operational behavior of the vehicle. The directive encompasses the motion system $\mathcal{S}_y$ along with a set of planning instructions $\mathcal{I}$, which collectively govern the planning process. This ensures that the actions of the vehicle align with human intentions, thereby maintaining both efficiency and strict adherence to operational constraints. The instruction $I_t$ can be expressed as follows:
\begin{equation}
I_t = \{\mathcal{S}_y,\mathcal{I}\}.
\end{equation}
We comprehensively detail the entire planning task in instruction $I_t$, by referencing the format established by GPT-Driver \cite{gpt_driver}. For the description of the motion system $\mathcal{S}_y$, we designed it as follows. Firstly, we elaborated on the critical coordinate systems for motion planning: the forward direction aligns with the positive Y-axis, the positive X-axis is perpendicular to the Y-axis on the right, and the yaw angle represents the counterclockwise rotation from the positive X-axis. Secondly, we defined the attributes of the vehicle. Each object (including the planned vehicle, surrounding vehicles, pedestrians, etc.) is represented as a quadrilateral to better illustrate potential collisions. Information about these objects includes their shape (coordinates of the four vertices of the quadrilateral), pose, velocity, and other kinematic parameters. Furthermore, we outlined the planning objectives, which include considerations of the current scenario, action planning, and trajectory generation. 

For the description of the planning instructions $\mathcal{I}$, we designed it as follows. Firstly, we have designed instructions to enable the planner to perform high-level semantic action representations. These include both preliminary coarse action planning, such as going straight, and refined final action planning, such as decelerating to go straight. Additionally, we have developed instructions for the planner to execute other human-like driving behaviors, such as collision avoidance, obeying traffic signals, and adhering to speed limits. 

For the input $I_p$, it encompasses a comprehensive set of descriptions of the planning inputs. Specifically, this input includes two primary components: the environmental observations, denoted as $\mathcal{O}$, and the current state of the planned vehicle, represented as $\mathcal{S}_t$. These components collectively furnish a detailed portrayal of the surrounding environment and the vehicle's motion status, which are pivotal for effective planning. The input $I_p$ can be expressed as follows:
\begin{equation}
I_p = \{\mathcal{O},\mathcal{S}_t\}.
\end{equation}
We have designed the observations $\mathcal{O}$ as follows. Firstly, it encompasses detailed information on nearby objects. This information primarily includes their categories (vehicles, bicycles, or pedestrians), current motion parameters, and predicted future positions, which are crucial for collision avoidance. Moreover, it encompasses detailed map information, primarily including traffic signal status, lane speed limits, current lane coordinates, and lane boundary coordinates, which are crucial for the comprehension and assessment of lane scenarios. For the ego-state $\mathcal{S}_t$, the primary components include its pose, velocity, acceleration, and past trajectory, which are essential for self-planning.

\subsection{Chain of Instructions}
\label{sec:chain_of_Instructions}
Through the aforementioned process, we obtain $X_t$. Additionally, it is necessary to perform prompt design on the output text to acquire $Y_t$. $Y_t$ incorporates the InstructChain $\mathcal{I}_c$ alongside the final planning trajectory $\mathcal{T}$. The InstructChain reflects the instructions $\mathcal{I}$, offering an interpretation of the resultant trajectory, thereby significantly enhancing the explainability of the planning process. The output text $Y_t$ can be expressed as follows:
\begin{equation}
Y_{t} = \{\mathcal{I}_c,  \mathcal{T} \}.
\end{equation}

The InstructChain $\mathcal{I}_c$ encompasses four primary steps. The first step involves the preliminary action plan for the current scenario, such as progressing along the current lane. The second step entails predicting potential collisions by estimating the future position of the ego vehicle and identifying objects that may pose a collision risk, providing their relative positions to facilitate subsequent action planning. Objects within a 3-meter radius of the ego vehicle warrant attention, whereas objects within 1.5 meters require immediate caution. The third step considers factors related to the traffic environment, including traffic signals, speed limits, and lane lines. For traffic signals, the colors green, yellow, and red respectively denote the permissions to proceed, prepare to stop and stop. An unknown state indicates that the signal is not currently applicable. Regarding speed limits, when the speed is within the limit, the current speed is deemed safe; when approaching the speed limit, further acceleration is not permitted; when exceeding the speed limit, immediate deceleration is indicated. Concerning lane lines, the primary consideration is the violation of lane boundaries. The fourth step involves integrating the information obtained from the previous three steps to produce a final action plan. Subsequently, the final action plan will be utilized to map the ultimate trajectory $\mathcal{T}$. The generated trajectory $\mathcal{T}$ consists of a series of poses, each characterized by $x$ coordinates, $y$ coordinates, and $\theta$ yaw angles, which can be represented as follows:
\begin{equation}
\mathcal{T} = \{(x_1, y_1, \theta_1), (x_2, y_2, \theta_2), \ldots, (x_n, y_n, \theta_n)\}.
\end{equation}
The output trajectories $\mathcal{T}$ can be integrated into the nuPlan simulation environment for rigorous evaluation of motion planning. Extensive and detailed testing ensures robustness and effectiveness. Through the aforementioned prompt design, data can be utilized to fine-tune the model, resulting in an autonomous driving motion planner that aligns with human driving habits.

\subsection{InstrucDriver}
\label{sec:instrucDriver}
Through the aforementioned process, data with instructions that align with human driving habits were obtained. Subsequently, we refer to the method of LLaMA2-Accessory \cite{zhang2023llamaadapter, gao2023llamaadapterv2} to fine-tune this data accordingly. For the input text $X_{t}$, it is processed via the preprocessing function $P$. The preprocessed text is subsequently transformed into encoded vectors $X$ through the embedding function $E$. This embedding function maps the text into a high-dimensional space, encapsulating the semantic information within a continuous vector representation. The aforementioned process can be delineated as follows:
\begin{equation}
X = E(P(X_{t})).
\end{equation}
After the text $X_{t}$ is represented as encoded vectors $X$, it is subsequently processed through the model $F_m$. The model $F_m$ incorporates Low-Rank Adaptation (LoRA) technology \cite{lora}, which is designed to enhance the efficiency of model fine-tuning by leveraging low-rank matrix approximations. The output of this process is the aforementioned logical logarithm $Y_{log}$ for the vocabulary. The delineation of this process is as follows:
\begin{equation}
Y_{log} = F_m(X, Att(X, A, B, b), wR(X)),
\end{equation}
where $Att$ represents the Attention module \cite{attention}, with $b$ denoting its associated bias term. The matrices $A$ and $B$ are employed as low-rank approximations within the LoRA technology, aiming to reduce both the number of parameters and the computational complexity. Furthermore, $R$ denotes the RMSnorm (Root Mean Square Layer Normalization) \cite{rmsnorm}, a normalization technique that contributes to the stabilization of the training process and enhances convergence. The variable $w$ represents the weight parameter of the RMSnorm output. The parameters $A$, $B$, $b$, and $w$ are fine-tuned to develop a planner that aligns with human driving habits.

\section{Experiment}
\label{sec:experiment}

\subsection{Dataset}
\label{sec:dataset}
The nuPlan dataset offers an extensive collection of data encompassing a wide range of traffic scenarios and driving conditions, providing researchers and developers with rich experimental resources. Each data set within the nuPlan dataset corresponds to a specific scenario, from which relevant information is extracted using feature builders. We utilized the same feature builder as planTF \cite{plantf} to gather data on the motion of the ego vehicle for the past 2 seconds and the next 8 seconds, with a time resolution of 0.1 seconds, and includes up to 32 nearby objects. It is important to note that only a subset of the collected dataset was employed for training purposes due to different time resolutions. Specifically, our experiments were conducted using a time resolution of 0.5 seconds. Regarding feature processing, we adhered to the settings described in planTF, where the position of the ego vehicle is designated as the origin (0,0), and all other coordinates are defined relative to this point. Similarly, the yaw angle of the ego vehicle is set to zero, and the yaw angles of all other objects are adjusted relative to the direction of the ego vehicle.

\subsection{Evaluation Metrics}
\label{sec:metrics}
We utilize the official evaluation metrics provided by nuPlan, comprising the open-loop score (OLS), nonreactive closed-loop score (NR-CLS), and reactive closed-loop score (R-CLS). The OLS is a composite indicator, incorporating calculations such as the Average Distance Error (ADE), Average Heading Error (AHE), Final Distance Error (FDE), Final Heading Error (FHE), and the Missing Rate, among others. The NR-CLS and R-CLS are both closed-loop scores and are also comprehensive indicators. Their computations involve assessments of the appropriateness of the driving area and direction, similarity in driving trajectories, and adherence to traffic regulations. The primary distinction between them is that R-CLS integrates background traffic control through an IDM \cite{idm} during simulations \cite{plantf}. All scores range from 0 to 100.

\begin{table}[t]
  \caption{
  Comparison of simulation results with state-of-the-arts on Test14-random and Test14-hard benchmarks. Numbers in bold represent the highest values within the indicators, while numbers underlined indicate the second-highest values. The evaluation results for other methods are sourced from planTF\cite{plantf}. 
  }
  \label{tab:comparison_experiment}
  \centering
  \setlength{\tabcolsep}{4.8pt}
  \begin{tabular}{ll|lll|lll}
  \toprule
  \multicolumn{2}{c}{Planners} & \multicolumn{3}{c}{Test14-random} & \multicolumn{3}{c}{Test14-hard} \\
    \midrule
    Type & Method & OLS & NR-CLS & R-CLS & OLS & NR-CLS & R-CLS\\
    \midrule
    \multirow{2}{*}{Rule-based} 
    & IDM \cite{idm} & 34.15 & 70.39 & 72.42 & 20.07 & 56.16 & 62.26 \\
    & PDM-Closed \cite{PDM} & 46.32 & \underline{90.05} & \textbf{91.64} & 26.43 & 65.07 & \underline{75.18} \\
    \midrule
    \multirow{5}{*}{Learning-based}
    & RasterModel \cite{RasterModel} & 62.93 & 69.66 & 67.54 & 52.40 & 49.47 & 52.16 \\
    & UrbanDriver \cite{UrbanDriver} & 82.44 & 63.27 & 61.02 & 76.90 & 51.54 & 49.07 \\
    & GC-PGP \cite{GC-PGP} & 77.33 & 55.99 & 51.39 & 73.78 & 43.22 & 39.63 \\
    & PDM-Open \cite{PDM} & 84.14 & 52.80 & 57.23 & 79.06 & 33.51 & 35.83 \\
    & PlanTF \cite{plantf} & \textbf{87.07} & 86.48 & 80.59 & \textbf{83.32} & \textbf{72.68} & 61.70 \\
    \midrule
    \multirow{2}{*}{Hybrid}    
    & GameFormer \cite{Gameformer} & 79.35 & 80.80 & 79.31 & 75.27 & \underline{66.59} & 68.83 \\
    & PDM-Hybrid \cite{PDM} & 82.21 & \textbf{90.20} & \underline{91.56} & 73.81 & 65.95 & \textbf{75.79} \\
    \midrule
    LLM-based & InstructDriver (Ours) & \underline{85.17} & 70.31 & 66.96 & \underline{81.12} & 57.37 & 52.95 \\
  \bottomrule
  \end{tabular}
  \vspace{-0.4cm}
\end{table}

\begin{figure}[tb]
  \centering
  \includegraphics[width=1.0\linewidth]{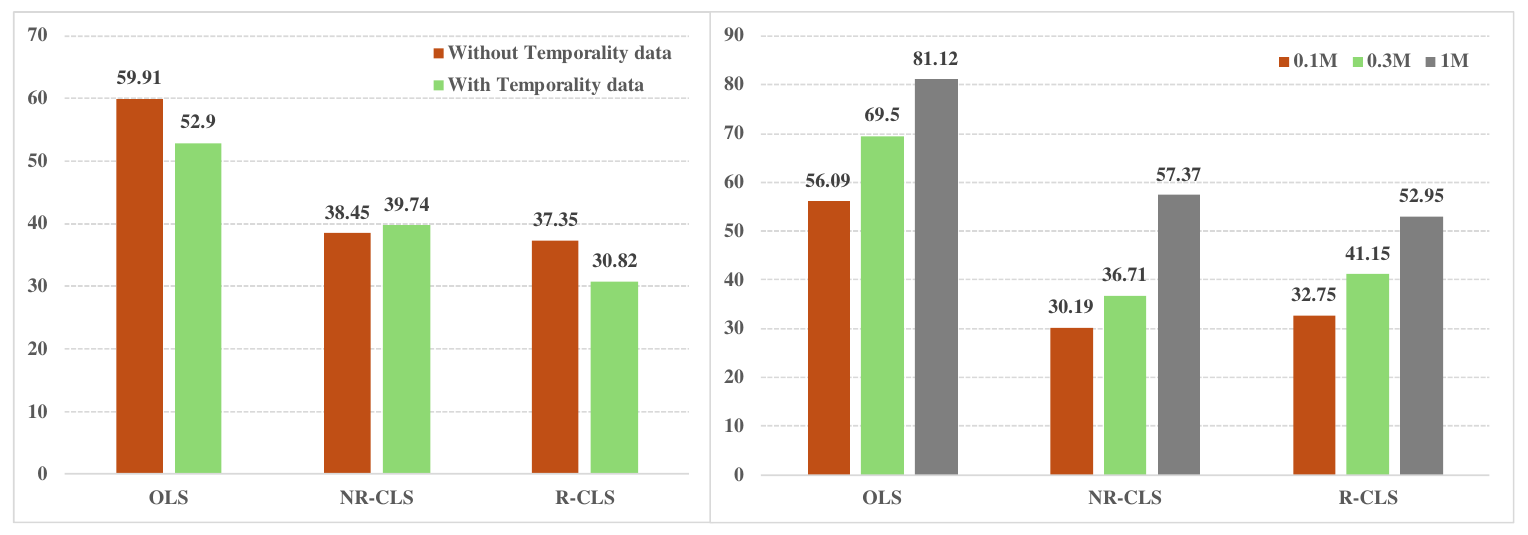}
  \vspace{-0.6cm}
  \caption
  {
  Comparison of simulation results with varying temporality and data volumes, with M denoting millions.
  }
  \label{fig:varying_data}
  \vspace{-0.4cm}
\end{figure}

\subsection{Implementation details}
\label{sec:implementation_details}
We utilized 8 NVIDIA 3090 GPUs for fine-tuning and a single NVIDIA 3090 GPU for inference. The batch size was set to 1, with each training session spanning 4 epochs and including a 1-epoch warmup phase. The learning rate was configured at 5e-5, with a minimum threshold of 5e-6. We set the weight decay at 0.02, applied gradient clipping at 2, and configured gradient accumulation to 2. For the input to the LLM, the maximum number of tokens was limited to 12,288. The temperature parameter was set to 0, and the top-p parameter was configured at 0.75.

\subsection{Main Results}
\label{sec:main_results}
We utilized a dataset of the same scale as that in planTF \cite{plantf}, comprising one million scenarios, to fine-tune LLaMA2-7B as our planner. Following the fine-tuning of the model, we evaluated its performance within the nuPlan simulation environment. This simulator uses a Linear Quadratic Regulator (LQR) controller for trajectory tracking and updates the state through an intrinsic kinematic model, based on control commands. The simulation parameters were set according to the specifications in \cite{plantf}, with a duration of 15 seconds and a frequency of 10 Hz. We conducted evaluations using Test14-random and Test14-hard, with the simulation results summarized in Table \ref{tab:comparison_experiment}. Our designed planner demonstrated commendable performance in both open-loop and closed-loop metrics, achieving the second-best result in open-loop metrics. Additionally, we visualized specific scenarios and their corresponding InstructChain, as depicted in Figure \ref{fig:planned_scenarios}. These visualizations indicate that our method produces human-like planning across various scenarios.

\subsection{Ablation and Analysis}
\label{sec:analysis}
\textbf{More scenarios are effective.} Handling temporally correlated data helps the model learn relationships between different time frames, leading to more coherent trajectory outputs. Exposure to diverse scenarios enhances the planner's contextual understanding and response strategies. We conducted comparative experiments to evaluate the impact of temporal correlations on model performance. One experiment used 300,000 scenarios without temporal correlations, while the other used 20,000 scenarios with 16 frames each. Despite the smaller data volume, the model trained without temporal correlations performed better, likely due to greater scenario diversity, as shown in the left image of Figure \ref{fig:varying_data}. Additionally, we compared models trained on varying data volumes. As depicted in the right image of Figure \ref{fig:varying_data}, results showed that exposure to more scenarios improves planning performance. Fine-tuning LLMs is computationally intensive, so controlling training data volume is crucial. These findings suggest that, when training data volume is fixed, prioritizing exposure to diverse scenarios is more important than enhancing trajectory coherence.

\begin{table}[tb]
  \caption{
  Comparison of simulation results with different model inputs. PT denotes the planned trajectory from the previous frame, while CC represents the corrected category, indicating that the category of the object is corrected based on its shape.
  }
  \label{tab:model_input_ablation}
  \centering
  \setlength{\tabcolsep}{16pt}
  \begin{tabular}{l|c|c|c|c|c}
    \toprule
    Model & PT & CC & OLS & NR-CLS & R-CLS \\
    \midrule
    InstructDriver-npt-ncc & & &   59.91 & 38.45 & 37.35   \\
    InstructDriver-pt-ncc & \checkmark &  &6.38 & 24.83 & 25.52   \\
    InstructDriver-npt-cc &  & \checkmark & 63.11 & 45.07 & 40.58   \\
  \bottomrule
  \end{tabular}
  \vspace{-0.3cm}

\end{table}

\begin{table}[tb]
  \caption{
  Ablation study of InstructChain. The steps s1, s2, s3, and s4 respectively represent the first, second, third, and fourth steps of the InstructChain process.
  }
  \label{tab:thought_chain_ablation}
  \centering
  \setlength{\tabcolsep}{11pt}
  \begin{tabular}{l|c|c|c|c|c|c|c}
    \toprule
    Model & s1 & s2 & s3 & s4 & OLS & NR-CLS & R-CLS \\
    \midrule
    no InstructChain & & & &    & 67.69  & 40.84  & 41.25  \\
    InstructChain-s1 & \checkmark & & &   & 68.10  & 39.49  & 40.88  \\
    InstructChain-s1-s2 & \checkmark & \checkmark & &    & 67.67  & 32.65  & 36.32  \\
    InstructChain-s1-s2-s3 & \checkmark & \checkmark & \checkmark &    & 67.22  & 35.15  & 38.23  \\
    InstructChain-full & \checkmark & \checkmark & \checkmark & \checkmark & 69.50 & 36.71 & 41.15 \\
  \bottomrule
  \end{tabular}
  \vspace{-0.5cm}
\end{table}

\textbf{Planned trajectory made planner learn only the mapping.} We attempted to use the planned trajectory from the previous frame as input for planning in the current frame, and the results are presented in Table \ref{tab:model_input_ablation}. It can be observed that incorporating the planned trajectory from the previous frame to input significantly reduces performance across all metrics. We contend that including the previous frame's planned trajectory leads the model to learn a simple mapping between the planned trajectories of successive frames, due to the minimal differences between them. Consequently, the trained planner disregards its own motion state and the perception of the surrounding environment, focusing solely on mapping the trajectory from the previous frame to the current one.

\textbf{Accurate input is crucial.} During our experiment, we observed that there were issues with reading object categories from the scenario. For instance, pedestrian data was sometimes categorized as unknown, leading to unexpected outputs from the planner. We addressed this issue by correcting the categories based on the size of the objects. As shown in Table \ref{tab:model_input_ablation}, the performance of the planner improved across all metrics after these category corrections. We believe that for methods involving LLMs, it is crucial to ensure the accuracy of even attributes with a relatively minor impact on planning, such as object categories, to achieve the expected outputs.

\begin{figure}[tb]
  \centering
  \includegraphics[width=1.0\linewidth]{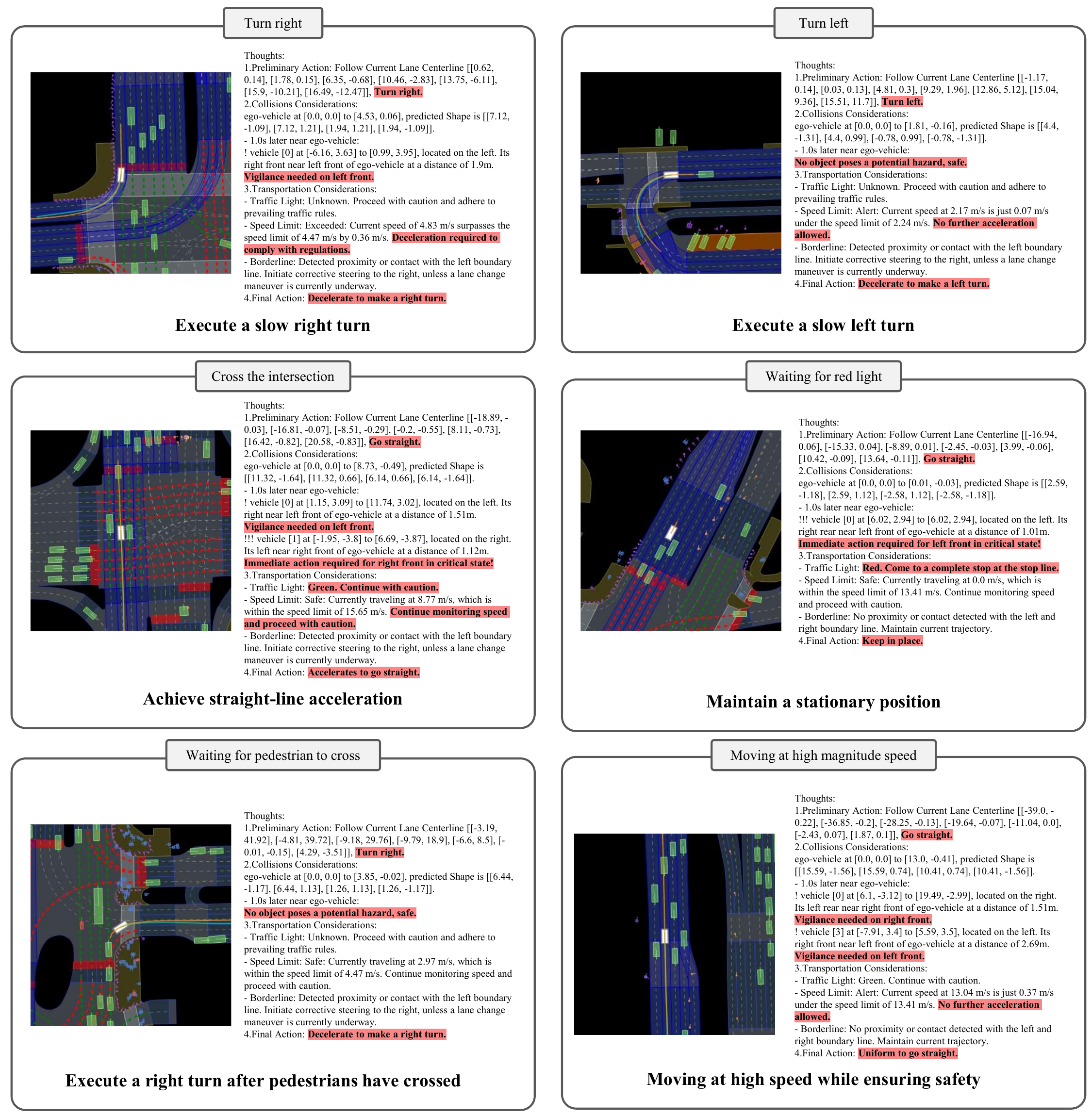}
  \vspace{-0.6cm}
  \caption
  {
  The illustration of the planning process, including specific scenarios and corresponding InstructChain, demonstrates that the planner can generate plans that align with human driving behavior based on given instructions.
  }
  \vspace{-0.5cm}
  \label{fig:planned_scenarios}
\end{figure}

\textbf{InstructChain makes planning more intuitive.} We conducted extensive ablation experiments on InstructChain, as illustrated in Table \ref{tab:thought_chain_ablation}. It can be observed that the inclusion of InstructChain enhances the open-loop metrics of the simulation results. More importantly, InstructChain provides a representation of the intermediate processes involved in output planning. Specifically, InstructChain reflects the entire planning process, including the understanding and reasoning based on the instructions. As shown in Figure \ref{fig:planned_scenarios}, using the example of crossing the intersection, InstructChain reveals that the planner initially proceeds straight based on the current lane markings, then notices collision risks from the left front and right front. Considering that the traffic light is currently green and there is no speeding, it finally plans to accelerate straight ahead. This indicates that the planner can perform motion planning consistent with human driving habits based on the given instructions.

\section{Limitations}
\label{sec:limitations}
The performance of InstructDriver still lags behind conventional methods. However, InstructDriver demonstrates aligned behaviors with humans through visualization analysis, demonstrating its potential to learn human-driving knowledge with instructions. The current use of LLMs for motion planning is impractical for real-time applications, necessitating the consideration of more lightweight models. Furthermore, the proposed method's performance in closed-loop simulation experiments remains suboptimal, indicating a need for further instruction design to enhance closed-loop performance. Lastly, due to the high computational resource demands of LLM inference, the current method has not been simulated within the val14 framework, which includes more diverse scenarios.

\section{Conclusion}
\label{sec:conclusion}
In this paper, we propose the InstructDriver, designing driving instruction data and an interpretable InstructChain based on human driving logic, and fine-tuning LLM as a motion planner. Furthermore, we evaluate the performance of InstructDriver through both open-loop and closed-loop experiments within the nuPlan simulation framework, achieving competitive results. We conducted several different comparative experiments on training data, demonstrating that a larger variety of scenarios in the training data leads to better planning outcomes for LLMs. Lastly, we conduct extensive ablation studies on InstructChain and visualize several example scenarios, illustrating that the planner can generate human-like driving behaviors based on instructions. InstructChain reflects the entire planning process, including the comprehension and reasoning of the instructions.

\textbf{Acknowledgements.} This work was supported by the National Natural Science Foundation of China under Grant 62373356 and the Open Projects Program of State Key Laboratory of Multimodal Artificial Intelligence Systems.

\bibliographystyle{plain}
\bibliography{ref}

\end{document}